\documentclass[10pt,journal,compsoc]{IEEEtran}
\ifCLASSOPTIONcompsoc
  \usepackage[nocompress]{cite}
\else
  \usepackage{cite}
\fi
\ifCLASSINFOpdf
  \usepackage[pdftex]{graphicx}
  \graphicspath{{../pdf/}{../jpeg/}}
  \DeclareGraphicsExtensions{.pdf,.jpeg,.png}
\else
  \usepackage[dvips]{graphicx}
  \graphicspath{{../eps/}}
  \DeclareGraphicsExtensions{.eps}
\fi
\usepackage{amsmath}
\usepackage{algorithmic}

\usepackage{array}

\ifCLASSOPTIONcompsoc
 \usepackage[caption=false,font=footnotesize,labelfont=sf,textfont=sf]{subfig}
\else
 \usepackage[caption=false,font=footnotesize]{subfig}
\fi
\usepackage{amssymb}
\usepackage{threeparttable}
\usepackage{multirow}
\usepackage{booktabs}

\newcommand{\tabincell}[2]{\begin{tabular}{@{}#1@{}}#2\end{tabular}} 

\begin{document}
%
\title{A Pseudo Label-wise Attention Network for Automatic ICD Coding}
%
%
%
%

\author{Yifan~Wu,
        Min~Zeng,
        Ying~Yu,
        Yaohang~Li,
        and~Min~Li
\IEEEcompsocitemizethanks{
\IEEEcompsocthanksitem Yifan Wu, Min Zeng, Ying Yu and Min Li are with the School of Computer Science and Engineering, Central South University, Changsha, P.R. China. \protect\\
E-mail: \{wudejian789, zengmin, yuying, limin\}@csu.edu.cn
\IEEEcompsocthanksitem Yaohang Li is with with the Department of Computer Science, Old Dominion University, Nor-folk, USA.\protect\\
E-mail: yaohang@cs.odu.edu}
\thanks{Manuscript received April 19, 2005; revised August 26, 2015.}}

%
%

\markboth{Journal of \LaTeX\ Class Files,~Vol.~14, No.~8, August~2015}%
{Shell \MakeLowercase{\textit{et al.}}: Bare Demo of IEEEtran.cls for Computer Society Journals}
%



\IEEEtitleabstractindextext{%
\begin{abstract}
Automatic International Classification of Diseases (ICD) coding is defined as a kind of text multi-label classification problem, which is difficult because the number of labels is very large and the distribution of labels is unbalanced. The label-wise attention mechanism is widely used in automatic ICD coding because it can assign weights to every word in full Electronic Medical Records (EMR) for different ICD codes. However, the label-wise attention mechanism is computational redundant and costly. In this paper, we propose a pseudo label-wise attention mechanism to tackle the problem. Instead of computing different attention modes for different ICD codes, the pseudo label-wise attention mechanism automatically merges similar ICD codes and computes only one attention mode for the similar ICD codes, which greatly compresses the number of attention modes and improves the predicted accuracy. In addition, we apply a more convenient and effective way to obtain the ICD vectors, and thus our model can predict new ICD codes by calculating the similarities between EMR vectors and ICD vectors. Extensive experiments show the superior performance of our model. On the public MIMIC-III dataset and private Xiangya dataset, our model achieves micro f1 of 0.583 and 0.806, respectively, which outperforms other competing models. Furthermore, we verify the ability of our model in predicting new ICD codes. The case study shows how pseudo label-wise attention works, and demonstrates the effectiveness of pseudo label-wise attention mechanism.
\end{abstract}

\begin{IEEEkeywords}
Automatic ICD coding, pseudo label-wise attention, text multi-label classification. 
\end{IEEEkeywords}}

\maketitle

\IEEEdisplaynontitleabstractindextext

%
\IEEEpeerreviewmaketitle

\IEEEraisesectionheading{\section{Introduction}\label{sec:introduction}}

%
%
%
%
\IEEEPARstart{T}{he} International Classification of Diseases (ICD) codes are widely used to measure diseases and health conditions, help medical reimbursement policies, and promote biomedical exchanges between regions \cite{SLEE1978The,PMID:3376487}. It has been reported that America spends about \$25 billion a year on manual ICD coding \cite{lang2007consultant}. Therefore, efficient automatic ICD coding methods are significant and urgently demanded. Automatic ICD coding extracts the key diseases and symptoms according to the discharge diagnosis of Electronic Medical Records (EMR), and then convert them into corresponding ICD codes. Essentially, it is a multi-label classification problem. The two challenges of automatic ICD coding are: 1) ICD-10 contains about 26,000 unique ICD codes, while an EMR mostly contains fewer than 10 ICD codes. But we need to calculate probabilities separately for each ICD code, which results in a huge amount of redundancy. 2) The frequency of most ICD codes is very low. Some ICD codes only have a few samples or even no samples for model training, which results in poor prediction accuracy in rare categories and even failure to predict new ICD codes (not appear in the training set). 

Existing deep learning-based methods for automatic ICD coding mainly include DeepLabeler \cite{2018Automated}, (DR-)CAML \cite{2018Explainable}, MVC-(R)LDA \cite{Sadoughi2018MedicalCP}, MultiResCNN \cite{li2020icd}, LAAT \cite{Vu2020ALA}, ZAGCNN \cite{Anthony2019Few}, WGAN-GP based automatic ICD coding \cite{2020Generalized}, etc. These models generally used the label-wise attention mechanism \cite{2018Explainable}, which requires assigning attention weights to every word in the full EMRs for different ICD codes. For example, MIMIC-III dataset \cite{JohnsonMIMIC} contains nearly 9,000 ICD codes in total, which means that we need to compute nearly 9,000 attention modes for each EMR. But do we really need to compute all these attention modes? An EMR usually consists of thousands of words and is labeled by just several ICD codes. It is obvious that label-wise attention has a huge computational redundancy and high cost. As we know, the classification of ICD has a hierarchical tree-like architecture, that is, ICD codes under the same branch (or with the same prefix) represent similar diseases, and they should pay similar attention to the contents of EMR. Specifically, in our study, assuming that the total number of ICD codes is $n$, we only need to calculate $m$ attention modes for an EMR ($m<<n$) and the $m$ attention modes are enough to describe the meaning of the EMR from various aspects. Since the $m$ attention modes do not correspond to the real ICD codes, we call it pseudo label-wise attention mechanism, which greatly reduces computational redundancy and memory consumption, and has better performance. In addition, for the prediction of rare and new ICD codes, ZAGCNN is inspired by the idea of zero-shot text classification \cite{Pushp2017TrainOT} and realized it by learning the relationship between EMR vectors and ICD vectors. It used a Graph Neural Network (GNN) to learn the hierarchical information on ICD tree and obtained the ICD vectors. However, from ICD-9 to ICD-11, the ICD codes are continuously expanded. When new ICD codes are added to the ICD tree, the neighbors of some nodes will be changed, which will make the ICD vectors learned by GNN changed, resulting in the instability of the results. On the basis of ZAGCNN, Song et al. \cite{2020Generalized} utilized Generative Adversarial Networks (GAN) to generate pseudo data for new ICDs to improve zero-shot ICD prediction. But the training of GAN is difficult and costly, which is not conducive to practical application. Therefore, we apply a more convenient and efficient model, BERT-based pre-trained language model, to obtain the ICD vectors with hierarchical information. The input of the model is composed of the ICD code itself and its title content. The code part can provide the hierarchical information, and the title part can provide the semantic information. By using the model, we can obtain the ICD vectors easier. The experimental results prove its effectiveness in predicting new ICD codes. 

In conclusion, the contributions of this paper are as follows: 

1) We propose a novel pseudo label-wise attention mechanism for multi-label classification, which only requires a small amount of attention modes to be calculated. Pseudo label-wise attention automatically compresses and merges the attention modes of similar labels. In other words, we compute only one attention mode for similar ICD codes, which reduces the number of parameters and achieves a better performance of the model. 

2) We propose a more convenient and effective way to represent the ICD vectors. We take the ICD codes themselves and the ICD titles as the descriptions of ICDs, and use a BERT-based pre-trained model to capture its semantic and hierarchical features as the ICD vectors. Compared with other methods, the pre-trained BERT-based model is easy-to-use and web-accessible, and we can compute all ICD vectors conveniently and efficiently. 

3) We demonstrate the effectiveness of our model. In the case of memory consumption is only 30\% of other models with label-wise attention, our model not only performs best on the standard benchmark datasets but also performs well on the prediction of new ICD codes. LD-PLAM can predict ICD codes very well, especially in predicting rare ICD codes, which improves R@10 by 37\%, 11.7\% than existing start-of-the-art model in few-shot ICD codes and zero-shot ICD codes, respectively. 

We call our model as LD-PLAM: a \textbf{L}abel \textbf{D}escription-based \textbf{P}seudo \textbf{L}abel-wise \textbf{A}ttentional \textbf{M}odel. LD-PLAM not only performs well on the standard benchmark datasets but it also outperforms other competing methods in predicting rare and new ICD codes.


 

\section{Related Work}

In the past decades, ICD codes are undertaken by coders, who assign an ICD code to medical record according to a doctor's clinical diagnosis record. Nowadays, with the accumulation of EMRs, more and more computational methods for automatic ICD coding have been proposed. Generally speaking, these methods can be divided into rule-based methods and learning-based methods. 

\noindent\textbf{Rule-based methods.}
Rule-based methods aim to develop a set of fuzzy matching methods according to the corresponding coding rules and carry out fuzzy matching for each discharge diagnosis \cite{0Mapping,farkas2008automatic}. The methods require doctors to follow certain rules when recording discharge diagnosis to ensure the quality of matching. There are two main problems with this approach: 1) Due to synonyms and personal writing habits, it is difficult to achieve ideal performances through a set of fixed matching rules. 2) It is difficult for different hospitals to reach a uniform set of EMR writing rules, which means the methods are not universally applicable, and different hospitals in different countries need different matching schemes. 

\noindent\textbf{Learning-based methods.}
Learning-based methods are mainly based on the modeling and feature extraction of a large number of prior data, from which the matching patterns between EMRs and ICD codes are learned. When there were few ICD labeling data, conventional machine learning-based methods for automatic ICD coding were mainly studied \cite{7559717,Perotte2014Diagnosis,2006Automating,2008From}. However, this kind of method is generally suitable for small sample learning. As the dataset scale increases, machine learning-based methods will gradually reach a bottleneck, while deep learning-based methods will not. Deep learning-based methods for automatic ICD coding are to use a deep neural network to learn a more desired semantic representation of the EMR \cite{Shi2017TowardsAI,2018Automatic,2019Automatic,2018Automated,Wu2020KAICD}, which has become the mainstream method nowadays. In this paper, we choose to build an efficient deep learning model for automatic ICD coding.

Multi-scale CNN \cite{2014Convolutional} is widely applied in deep learning-based methods to capture the key information of EMRs. On this basis, DeepLabeler \cite{2018Automatic} combined clinic notes' global features extracted by doc2vec \cite{pmlr-v32-le14} to infer ICD codes. After that, KAICD \cite{Wu2020KAICD} introduced the semantic information of ICD titles to build a knowledge database. For different EMR inputs, KAICD learned a knowledge vector, and thus the EMR vectors computed by multi-scale CNN and the knowledge vectors were concatenated to predict ICD codes. Considering the differences between ICD codes, CAML \cite{2018Explainable} applied the label-wise attention to automatic ICD coding. Moreover, DR-CAML added the description information of ICD titles to the loss function as a regular term to improve the performance on predicting rare ICD codes. MVC-LDA \cite{Sadoughi2018MedicalCP} used the multi-view CNN as feature extractor, and also used the label-wise attention mechanism. Additionally, it applied the shift bias strategy to learn different thresholds in the prediction based on the occurrence of different ICD codes. MVC-LDA also added the same regular term as DR-CAML into the loss function, named MVC-RLDA. After that, Li et al. \cite{li2020icd} noted that the length and grammar of text fragments vary a lot in different EMRs and thus they applied a multi-filter residual CNN to capture various text patterns with different lengths. 


\begin{figure*}
    \centering
    \includegraphics[scale=0.65, trim=0 0 0 0]{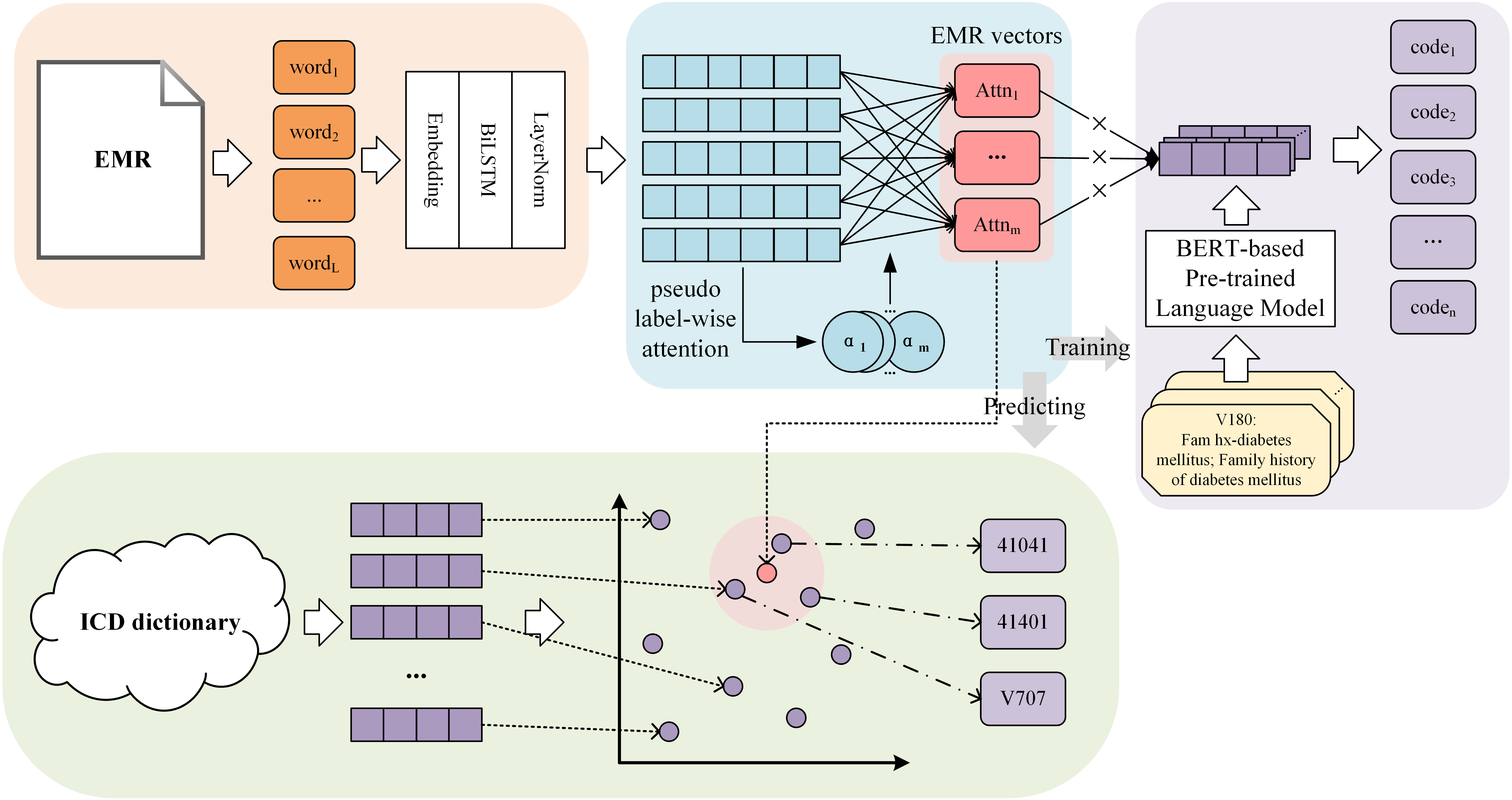}
    \caption{The architecture of LD-PLAM. The input of LD-PLAM is an EMR. First, we encode the words in EMR as vectors. After embedding words, we use a bidirectional LSTM and the pseudo label-wise attention to obtain the feature vectors of an EMR. Then, we extract the descriptions of every ICD code in dataset and then a BERT-based pre-trained language model to compute the ICD vectors. Finally, we calculate the probability of each ICD code based on the similarity between EMR vectors and ICD vectors. }
    \label{fig:01}
\end{figure*}

\section{Problem statement and notations}
Given a clinic note, our task is to design a model to predict the associated ICD codes. As with a general Natural Language Processing (NLP) task, we think of the clinic note as a sequence of words $w = \begin{pmatrix} w_1 & w_2 & ... & w_{l_r} \end{pmatrix}$ that serve as input to the model, where $l_r$ is the length of a clinic note. At the same time, each ICD code is regarded as a predicted label of the model. In this way, for each clinic note $w$, our prediction goal is to get a 0/1 label vector $y = \begin{pmatrix} y_1 & y_2 & ... & y_n \end{pmatrix}$, where $n$ represents the total number of ICD codes, $y_i=0$ means there is no ICD code $i$, and $y_i=1$ means there is ICD code $i$. Obviously, this a multi-label classification problem composed of many binary classification tasks, and we will use the binary cross entropy function as the optimization objective. The binary cross entropy for multi-label classification is defined as: 

\begin{equation}
L(y^\prime, y) = \sum_{i=1}^{n} y_i\mathrm{log}{y^\prime_i} + (1-y_i)\mathrm{log}(1-y^\prime_i)
\end{equation}

\noindent where $n$ is the total number of ICD codes, $y^\prime_i \in [0,1]$ is the predicted value of ICD code $i$, $y_i \in \{0,1\}$ is the true valud of ICD code $i$. In addition, for the sake of convenience, we summarize some important notations used throughout the paper in Table \ref{tab:6}. 

\begin{table}
\caption{Notations definition}
\centering
\begin{tabular}{cp{6.5cm}l}
\toprule
Symbol                                            & Definition or Description\\
\midrule
$n$                                               & the number of ICD codes \\
$m$                                               & the number of pseudo labels, $m<<n$ \\
$l_r$                                             & the length of EMRs \\
$l_d$                                             & the length of ICD description \\
$d_c$                                             & the number of hidden units in bidirectional LSTM \\
$x$                                               & the output of bidirectional LSTM \\  
$\alpha$                                          & the attention score of words in pseudo labels \\  
$e$                                               & the word vector obtained by RoBERTa \\
$u$                                               & the attention vector obtained by pseudo label-wise attention \\
$v$                                               & the ICD code vector obtained by computing the average of the word vectors $u$ \\
$w,b$                                             & the weights and bias terms in feed forward neural network \\

\bottomrule
\end{tabular}
\label{tab:6}
\end{table}

\section{Methodology}
As shown in Figure \ref{fig:01}, for representing an EMR, LD-PLAM uses a bidirectional LSTM and the pseudo label-wise attention to get EMR vectors; for representing a ICD code, a BERT-based pre-trained language model is applied to calculate ICD vectors. Finally, LD-PLAM predicts the results based on the relationship between the EMR vectors and the ICD vectors. For the new ICD codes, we only need to calculate the ICD vectors in the same way to perform predictions. 

\subsection{Representation of EMRs}
In order to estimate the similarity between ICD vectors and EMR vectors, the first thing is to explore how to obtain the representation of EMR vectors. We preprocess the original EMRs, and then get the final EMR vectors through a bidirectional Long Short-Term Memory (LSTM) network and the pseudo label-wise attention. 

\noindent\textbf{EMRs preprocessing.}
The original sequence length of EMR is very large and inconsistent. In order to facilitate the calculation, we need to unify the length of EMRs. Usually, a maximum length will be set: fill the EMRs that are less than the length, and intercept the EMRs that are over the length. Considering that some useful information might be lost by direct truncation, for those EMRs that are too long, we calculate the importance of each word in EMRs separately. After that, the words are ranked in order of importance, and the most unimportant words are removed from the EMR to reduce the sequence length. We use Term Frequency-Inverse Document Frequency (TF-IDF) to measure the importance of a word, which is defined as the product of term frequency and inverse document frequency. 





\noindent\textbf{Bidirectional LSTM network.}
After preprocessing, an embedding layer is used to vectorize each word in EMRs, and it will be fine-tuned during the training process. We use skip-gram algorithm \cite{Mikolov2013EfficientEO} to perform pre-training on all EMRs and initialize the embedding layer.  Then, a bidirectional LSTM is applied to extract key information in EMRs. It can extract key phrases of different lengths by using the gate units to weaken or enhance the word features in two directions, which is very similar to ICD coders' extraction of key diagnostic information or symptom description from EMRs. Last, we use layer normalization \cite{Ba2016LayerN} to prevent overfitting.  

\noindent\textbf{Pseudo label-wise attention.}
After the bidirectional LSTM, a sequence vector is obtained. Note that different ICD codes vary greatly, which means their corresponding key phrases are not consistent. In other words, we need to focus on the sequence vectors in different attention modes. Assuming $n$ is the number of ICD codes, label-wise attention needs to calculate $n$ attention modes for each word in the sequence, and each attention mode corresponds to an ICD code. Label-wise attention plays a great role in multi-label prediction tasks. (DR-)CAML \cite{2018Explainable}, MVC-(R)LDA \cite{Sadoughi2018MedicalCP}, etc. all applied this attention structure. However, do we really need to compute different attention modes for every different ICD codes? As mentioned above, the ICD codes with a same prefix represent a same category or subcategory of diseases and they should have similar attention modes. Therefore, we can only calculate fewer attention modes, which is enough to express an EMR from the perspective of different ICD categories. Since the fewer attention modes do not correspond to the real ICD codes, we call it pseudo label-wise attention mechanism. Given $x \in \mathbb{R}^{l_r \times d_c}$ which represents the output after the bidirectional LSTM, the attention vector $u_i \in \mathbb{R}^{1 \times d_c}$ of pseudo label $i$ is computed as:

\begin{equation}
u_i = \mathrm{h}( \sum_{j=1}^{l_r} \alpha_{ij} x_j )
\end{equation}

\noindent where $l_r$ is the length of an EMR, $d_c$ is the number of hidden units in the bidirectional LSTM, function $\mathrm{h(\cdot)}$ is a nonlinear layer with $\mathrm{ReLU(\cdot)}$ activation whose dimension of output is the same as ICD vectors, $x_j \in \mathbb{R}^{1 \times d_c}$ is the $j$-th row of $x$, $\alpha_{ij}$ is the score of word $j$ in pseudo label $i$ and $\alpha_{i} \in \mathbb{R}^{1 \times l_r}$ is defined as:

\begin{equation}
\alpha_{i} = \begin{pmatrix} \alpha_{i1} & \alpha_{i2} & ... & \alpha_{il_r} \end{pmatrix} = \mathrm{softmax}(xw_i+b_i)
\end{equation}

\noindent among them, $w_i \in \mathbb{R}^{d_c \times l_r}$ and $b_i \in \mathbb{R}^{1 \times l_r}$ are learnable parameters. Each attention mode $\alpha_i$ corresponds to an output: the attention vector $\alpha_i$. In total, we learn $m$ attention modes $\alpha_1, \alpha_2, ..., \alpha_m$ to get $m$ attention vectors $u_1, u_2, ..., u_m$ ($m<<n$) and the $m$ attention vectors together constitute the representation of an EMR. 

Furthermore, we can also carry out multi-layer pseudo label-wise attention to capture deeper features. Set $a_i^{(k)}$ is the $k$-th layer's attention vector, and it can be computed as:

\begin{equation}
a_i^{(k)} = \sum_{j=1}^{l_r^{(k-1)}} \alpha_j^{(k-1)} u_j^{(k-1)}
\end{equation}

\begin{equation}
u_j^{(k-1)} = \mathrm{h^{(k-1)}}(a_j^{(k-1)})
\end{equation}

\begin{equation}
\alpha_i^{(k-1)} = \mathrm{softmax}(a^{(k-1)}w_i^{(k-1)}+b_i^{(k-1)})
\end{equation}

\noindent where $l_r^{(k-1)}$ represents the number of pseudo ICD categories in $(k-1)$-th layer, function $\mathrm{h^{(k-1)}(\cdot)}$ is a nonlinear layer with $\mathrm{ReLU(\cdot)}$ activation whose dimensions of input and output are the same, $w_i^{(k-1)} \in \mathbb{R}^{d_c \times l_r^{(k-1)}}$ and $b_i^{(k-1)} \in \mathbb{R}^{1 \times l_r^{(k-1)}}$ are the learnable parameters in $(k-1)$-th attention layer. We will take the last layer's $u_1, u_2, ..., u_m$ as the final EMR vectors. 

\subsection{Representation of ICD Codes}
In addition to computing the EMR vectors, we also need to vectorize the ICD codes. Our idea is to extract the language description of each ICD, and use a pre-trained language model to encode the sentence.  

\noindent\textbf{Composition of ICD description.}
We mainly use two parts of information to compose the ICD description: ICD codes itself and ICD titles. The ICD codes, as a tree-like disease code composed of letters and numbers, inherently contain hierarchical information. For example, the code "H53.011" (from ICD-10-CM \cite{Plummer2015International}): "H53" means the disease category is "visual disturbances"; "H53.01" means the disease subcategory is "deprivation amblyopia", and it is a specific type under visual disturbances; "H53.011" means the disease code is "deprivation amblyopia, right eye", and it specifies the part of "deprivation amblyopia" is the right eye. In other words, the ICD codes with the same prefix will be under the same branch, representing the same class of diseases. Therefore, we can use the ICD codes itself as a part of the ICD description to capture the hierarchy of ICD codes. In addition, the ICD titles contain the description of the diseases corresponding to its ICD codes, which is obviously an important component of the ICD description. As a result, we also use ICD titles to compose the ICD description so as to better construct the semantic relationship between the ICD vectors and the EMR vectors. The final ICD description can be expressed as "\emph{xxxxxx (ICD code): ...... (ICD title)}". 

\noindent\textbf{BERT-based pre-trained language model.}
After extracting the ICD description, we need to vectorize them through a pre-trained language model. BERT is a large-scale pre-trained language model based on transformer, which has refreshed the records of many NLP tasks \cite{Devlin2019BERTPO}. This kind of pre-trained language model can well encode the ICD description and get an efficient context-dependent ICD vector. RoBERTa, as an improved version of BERT, further improves the performance by removing the next sentence prediction object and applying the dynamic mask mechanism to train on a larger corpus \cite{Liu2019RoBERTaAR}. For better performances, we use the RoBERTa as our pre-trained language model to encode each ICD description separately, and average the word vectors as the representation of ICD codes. We define $v_{i}$ as the vector of ICD $i$, $e_{ij}$ as the word $j$'s vector in the description of ICD $i$:

\begin{equation}
v_{i} = \frac{1}{l_{d}} \sum_{j=1}^{l_{d}} e_{ij}
\end{equation}

\noindent where $l_{d}$ represents the length of ICD description. Thus, as long as the ICD code has its ICD title, the ICD vector can be computed in this way. 

\subsection{Calculation of the Similarity between EMR Vectors and ICD Vectors}
Finally, we need to calculate the similarity between EMR vectors and ICD vectors. During training, the similarity between EMR vectors and their positive ICD vectors will be maximized, while the similarity between EMR vectors and their negative ICD vectors will be minimized. During prediction, the occurrence probability of each ICD code can be inferred by the similarity. 

Based on the previous calculation, the $m$ attention vectors $u_1, u_2, ..., u_m$ are obtained for each EMR, and ICD code $i$'s vector $v_i$ is computed. To calculate the similarity, the function $\mathrm{h(\cdot)}$ transforms $u_1, u_2, ..., u_m$ to the same dimension as $v_i$. The similarity $s_{u,v_i} \in \mathbb{R}$ between the EMR vectors $u = \begin{pmatrix} u_1 & u_2 & ... & u_m \end{pmatrix} ^\mathrm{T} \in \mathbb{R}^{m \times d_t}$ and the ICD vector $v_i \in \mathbb{R}^{1 \times d_t}$ are defined as: 

\begin{equation}
s_{u,v_i} = \mathrm{f}(\sum_{j=1}^{m} \beta_{ij} u_j)
\end{equation}

\begin{equation}
\beta_{ij} = \mathrm{softmax}(u_j v_i^\mathrm{T})
\end{equation}

\noindent where function $\mathrm{f(\cdot)}$ is a nonlinear dense layer with $\mathrm{sigmoid(\cdot)}$ activation whose dimension of output is $1$, $d_t$ is the length of ICD vectors, $\beta_{ij} \in \mathbb{R}$ represents the simiarity between $u_j$ and $v_i$. Finally, we can compute the binary cross entropy based on the predicted value $s_{u,v_i}$ and the true value, and optimize the model. 

\subsection{Comparison between pseudo label-wise attention and label-wise attention}
\begin{figure}
    \centering
    \includegraphics[scale=0.5, trim=0 0 0 0]{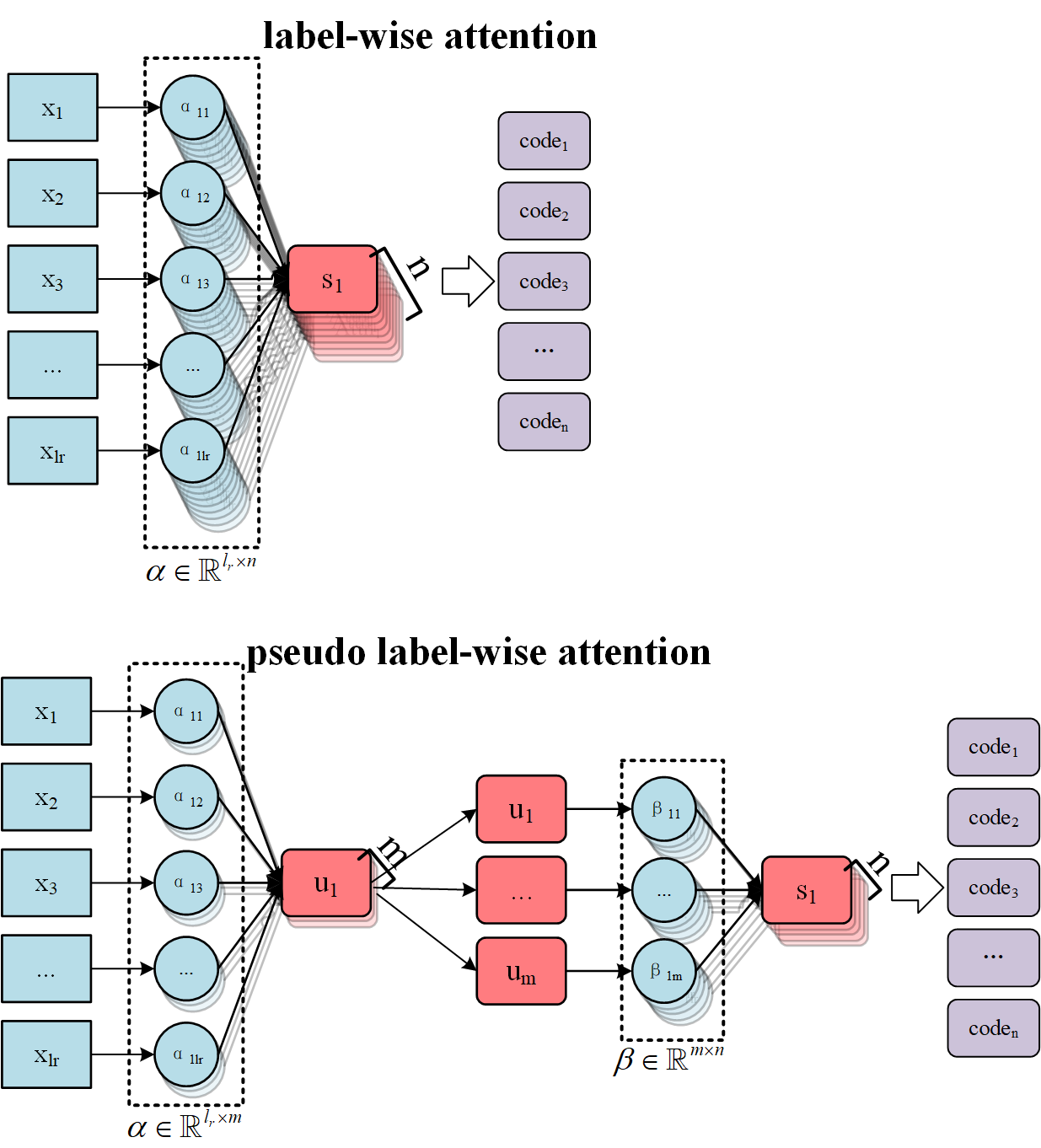}
    \caption{Comparison between pseudo label-wise attention and label-wise attention. }
    \label{fig:06}
\end{figure}

In general, the core difference between our approach and others is shown in Figure \ref{fig:06}. Assume that $x=\{x_1,x_2,...,x_{l_r}\} \in \mathbb{R}^{l_r \times d_c}$ represents the semantic vectors of words extracted from the clinic notes, and $\alpha, \beta$ represent the attention scores of each word under each ICD code or pseudo label. In the case of ignoring the activation function, for label-wise attention, we compute the final multi-label classification results by matrix multiplication $x^T\alpha$, which requires $d_c \times l_r \times n$ multiplications and a $l_r \times n$ matrix to be saved. And for pseudo label-wise attention, the final multi-label classification results are obtained by matrix multiplication $x^T\alpha\beta$, which needs $d_c \times l_r \times m + d_c \times m \times n$ multiplications and matrices of $l_r \times m, m \times n$ to be saved. Here, $l_r$ represents the length of an EMR, which is generally reserved 2000-3000 words; $d_c$ is the number of hidden layer neurons of CNN or RNN, which is generally 64-256 units. When the label number $n$ is very large, the pseudo label-wise attention will save a lot of memory than the label-wise attention. For example, if $l_r=2500, n=10000, m=128, d_c=128$, then label-wise attention requires $3200 M$ multiplications and $25 M$ elements to be saved, while pseudo label-wise attention only needs $204.8 M$ and $1.6 M$ elements to be saved ($1M=10^6$). Additionally, as we know, deep models have better generalization than wide models \cite{cheng2016wide}, and the pseudo label-wise attention does just that. To sum up, pseudo label-wise attention not only reduces the number of parameters, but also increases the number of layers of the attention modules. And this is why pseudo label-wise attention is better than label-wise attention. 

\section{Experiments}

\subsection{Datasets}
In order to demonstrate the effectiveness of LD-PLAM, we choose two real-world dataset about ICD coding to show LD-PLAM's performance. MIMIC-III (a public dataset in English) and Xiangya (a private dataset in Chinese) datasets are used in the experiments. The statistical details of the two datasets are shown in Table \ref{tab:5}

\begin{table}
\caption{The details of datasets}
\centering
\begin{tabular}{ccc}
\toprule
                            & MIMIC-III & Xiangya     \\
\midrule
Total \# of EMRs             & 52,726   & 7,732      \\
Total \# of codes            & 8908     & 1,177      \\
Avg \# of codes per EMR      & 15.9     & 3.6        \\
Max \# of codes per EMR      & 71       & 14         \\
Std \# of codes per EMR      & 8.1      & 2.1        \\
\bottomrule
\end{tabular}
\label{tab:5}
\end{table}

\noindent\textbf{MIMIC-III.}
MIMIC-III is a public dataset of medical information for intensive care that contains more than 58,000 hospitalizations of 38,645 adults and 7,875 newborns at Beth Israel Deaconess Medical Center from June 2001 to October 2012. We extract 52,726 discharge summaries and their ICD-9 codes. In order to ensure the comparability of the results, we follow the same setting and division with \cite{2018Explainable}, and finally get two datasets: MIMIC-III full and MIMIC-III 50. MIMIC-III full needs to predict all nearly 9000 ICD codes, while MIMIC-III 50 only needs to predict the 50 ICD codes with the highest frequency. 

\noindent\textbf{Xiangya.}
Xiangya dataset is a Chinese private dataset. It has gathered 7,732 EMRs and their ICD-10 codes from three affiliated hospitals in recent years. There are a total of 1,177 ICD-10 codes, with an average of 3.6 codes per EMR. Same as MIMIC-III, we prepare two experimental datasets: Xiangya full and Xiangya 50. Among them, we randomly divide the dataset, using 80\% for training, 10\% for validation, and 10\% for testing. Meanwhile, we ensured that the patient IDs of training set, validation set and testing set are not the same. 

\subsection{Baselines}
In experiments, LD-PLAM is compared against the previous state-of-the-art methods on automatic ICD coding. For general experimental comparisons, we choose DeepLabeler \cite{2018Automated}, KAICD \cite{Wu2020KAICD}, (DR-)CAML \cite{2018Explainable}, MVC-(R)LDA \cite{Sadoughi2018MedicalCP}, MultiResCNN \cite{li2020icd} as the baselines$^1$. For the experiments of predicting fare ICD codes, ESZSL \cite{romera2015embarrassingly} and ZAGCNN \cite{Anthony2019Few} are compared. Some of these models are reproduced in our code framework, and we follow the same parameter settings as their original papers. 

\footnotetext[1]{As for LAAT \cite{Vu2020ALA}, we don't choose it for comparison because we observed that in MIMIC-III dataset, clinic notes with different retained length would have some influence on the results. LAAT retains 4000 words for each clinic note, 1500 words more than the 2500 words in other models. This may make the comparison unfair. }

\noindent\textbf{DeepLabeler.} DeepLabeler predict ICD codes by combining local features and global features of clinic notes. The local features are captured by a multi-scale CNN with max pooling. The global features are extracted by a doc2vec \cite{pmlr-v32-le14} model. 

\noindent\textbf{KAICD.} KAICD uses the semantic information of ICD titles to build a knowledge database. For different EMR inputs, a knowledge vector is obtained through knowledge attention, and the EMR vectors computed by multi-scale CNN and the knowledge vectors are concatenated to perform the prediction of ICD codes. In the original paper, Wu et al. encoded each word as a 320-dimensional vector, and 448 convolution kernels were used in their multi-scale CNN. 

\noindent\textbf{(DR-)CAML.} The main idea of CAML is to apply the label-wise attention to automatic ICD coding. In addition, they further proposed DR-CAML, which adds the description information of ICD titles to the loss function as a regular term to improve the performance on predicting the rare ICD codes.  

\noindent\textbf{MVC-(R)LDA.} MVC-LDA uses the multi-view CNN as their feature extractor, and also uses the label-wise attention mechanism. Additionally, MVC-LDA applies the shift bias strategy to learn different thresholds in the prediction function based on the occurrence of different ICD codes. They added the same regular term based on ICD title description as DR-CAML into the loss function, namely MVC-RLDA. 

\noindent\textbf{MultiResCNN.} MultiResCNN captures various text patterns with different length by their multi-filter and residual convolutional layer. 

\noindent\textbf{ESZSL.} ESZSL is a very simple method to achieve zero-shot learning, which consists of only two layers of neural network to learn the relationship between two vectors. 

\noindent\textbf{ZAGCNN.} Their main idea is to use a GNN to extract the hierarchy features from the ICD tree to obtain the ICD vectors. They realized the prediction of few-shot ICD codes and zero-shot ICD codes on learning the relationship between the EMR vectors computed by GNN and the ICD vectors. 

\subsection{Training Details}
We find that a single layer of pseudo label-wise attention is sufficient to achieve a good result in automatic ICD coding, so in our experiments, we only use one-layer pseudo label-wise attention. We conduct a large number of experiments on our device with GPU of NVIDIA V100 32GB to fine tune the hyper-parameters. Our final parameter settings are shown in Table \ref{tab:0}. For MIMIC-III and Xiangya datasets, we encode each word as a 128-dimensional vector that is pre-trained using skip-gram algorithm. Also, we use dropout with rate 0.2 after word embedding. We retain 2500 words for every EMRs based on the TF-IDF value (note that we treat each character as a token, meaning that we retain 2500 characters for Xiangya's EMRs). We use 128 hidden units in the bidirectional LSTM layer for both datasets. For MIMIC-III full dataset with 8,907 ICD codes and Xiangya full dataset with 1,177 ICD codes, we set the number of pseudo labels to 256 and 64, respectively. For MIMIC-III 50 and Xiangya 50 datasets, we set the number of pseudo labels to 16. During the training process, we use LAMB \cite{You2020LargeBO} optimizer with learning rate 0.005 and the training process is stopped when the metrics of validation set has not been improved for more than 64 rounds. Considering the limitations of hardware, the batch size is set to 128 for MIMIC-III full and Xiangya full dataset, 256 for MIMIC-III 50 and Xiangya 50 dataset. 

As for the baselines: for MIMIC-III dataset, most of the results are from their papers; for Xiangya dataset, we retrain these models using their optimal parameter settings and obtain the results. 

\begin{table*}[ht]
\caption{hyper-parameters setting of LD-PLAM. }
\centering
\setlength{\tabcolsep}{1mm}{
\begin{tabular}{ccccccccccccc}
\toprule
hyper-parameters & MIMIC-III full & MIMIC-III 50 & Xiangya full & Xiangya 50 \\
\midrule
embedding size   &  128  & 128 & 128 & 128 \\
number of LSTM units      &  128  & 128 & 128 & 128 \\
number of pseudo labels   &  256  &  16 & 48 &  16 \\
dropout rate     &  0.2  & 0.2 & 0.2 & 0.2 \\
threshold        &  0.5  & 0.5 & 0.5 & 0.5 \\
batch size       &  128  & 256 & 128 & 256 \\
learning rate    & 0.005 & 0.005 & 0.005 & 0.005 \\
\bottomrule
\end{tabular}
}
\label{tab:0}
\end{table*}

\subsection{Comparison with Other Methods}
Experiments are carried out on MIMIC-III 50, MIMIC-III full, Xiangya 50 and Xiangya full, respectively. The results of micro F1 (MiF), micro AUC (MiAUC), precision at 5 (P@5), precision at 8 (P@8) are reported on the test set, as shown in Table \ref{tab:1} and Table \ref{tab:2}. P@$n$ is defined as the proportion of real labels in the most possible $n$ labels predicted by the model. Following previous studies, for full datasets with more ICD categories, we choose P@8 as the metrics; for 50 datasets with fewer ICD categories, we choose P@5 as the metrics. 

\noindent\textbf{Comparison on MIMIC-III dataset.} As shown in Table \ref{tab:1}, for MIMIC-III full dataset, models like DeepLabeler or KAICD without label-wise attention performs poorly, with MiF lower than 0.5, while (DR-)CAML, MVC-(R)LDA, MultiResCNN, LD-PLAM with label-wise attention generally achieves MiF over 0.5, indicating that label-wise attention is very effective in multi-label tasks with a large number of labels. Using pseudo label-wise attention, LD-PLAM only computes 256 attention modes, which is 97.1\% less than 8,907 attention modes of general label-wise attention. In addition, LD-PLAM achieves the best MiF, MiAUC, P@8 of 0.583, 0.986, 0.756, respectively. And the memory cost of LD-PLAM is significantly lower than that of other models with label-wise attention, due to the pseudo label-wise attention. For MIMIC-III 50 dataset, the label-wise attention also shows obvious advantages. Due to its large dimension of word embedding and the number of convolution kernels, DeepLabeler and KAICD also achieves a great MiF of 0.634 and 0.640, even exceeding the (DR-)CAML. The results indicate that it is beneficial to increase the dimension of word embeddings and the number of convolution kernels in multi-label classification tasks with only a few labels. LD-PLAM, computing only 16 attention modes, which is 68\% less than (DR-)CAML, MVC-(R)LDA and MultiResCNN, achieves the best results on all indicators, with MiF of 0.699, MiAUC of 0.937, and P@5 of 0.660. We speculate that this is due to the high-quality ICD vectors and the powerful bidirectional LSTM. 

\begin{table}[ht]
\caption{Comparison of LD-PLAM and other competing models on MIMIC-III dataset. }
\centering
\setlength{\tabcolsep}{1mm}{
\begin{tabular}{ccccccccccccc}
\toprule
\multirow{2}{*}{Models} & \multicolumn{3}{c}{MIMIC-III full} &\multicolumn{3}{c}{MIMIC-III 50}\\
\cmidrule(lr){2-4}\cmidrule(lr){5-7}
                        &   MiF   &  MiAUC  &   P@8    &    MiF  &  MiAUC  &   P@5    \\
\midrule
DeepLabeler             &  0.401  &  0.976  &  0.567   &  0.634  &  0.938  &  0.623   \\
KAICD                   &  0.434  &  0.973  &  0.603   &  0.640  &  0.925  &  0.625   \\
CAML                    &  0.539  &  0.986  &  0.709   &  0.614  &  0.909  &  0.609   \\
DR-CAML                 &  0.529  &  0.985  &  0.690   &  0.633  &  0.916  &  0.618   \\
MVC-LDA                 &  0.543  &    -    &  0.705   &  0.668  &    -    &  0.644   \\
MVC-RLDA                &  0.550  &    -    &  0.712   &  0.674  &    -    &  0.641   \\
MultiResCNN             &  0.552  &  0.986  &  0.734   &  0.670  &  0.928  &  0.641   \\
\midrule
LD-PLAM                 &  \textbf{0.583}  &  \textbf{0.986}  &  \textbf{0.756}   &  \textbf{0.699}  &  \textbf{0.937}  &  \textbf{0.660}   \\
\bottomrule
\end{tabular}
}
\label{tab:1}
\end{table}

\noindent\textbf{Comparison on Xiangya dataset.} As shown in Table \ref{tab:2}, similar to MIMIC-III dataset, DeepLabeler and KAICD are generally worse than other models in Xiangya full dataset, while the difference is smaller on Xiangya 50, which further confirms that the label-wise attention plays an important role when there are a large number of labels. In addition, the relative performance of the models is changed greatly: on Xiangya full and Xiangya 50 datasets, the MiF of CAML reaches 0.730 and 0.803, respectively, which are higher than that of MVC-LDA (0.711 and 0.798). We speculate that this is because the shift bias strategy used by MVC-(R)LDA is not suitable for Chinese Xiangya data. Moreover, the addition of label description regularization reduces the performance of the model, which results in the MiF value decreases from 0.730 of CAML to 0.709 of DR-CAML, and from 0.711 of MVC-LDA to 0.697 of MVC-RLDA. Not surprisingly, LD-PLAM achieves the best results, and is significantly better than other models. On the one hand, this is due to the effect of the pseudo label-wise attention. On the other hand, this is because the optimal convolution window sizes obtained by (DR-)CAML, MVC-(R)LDA, MultiResCNN in English MIMIC-III dataset are no longer applicable to Chinese Xiangya data. Among them, MultiResCNN can capture key phases of different lengths to a certain extent because it has a larger receptive field through the multi-filter CNN stacked with residuals. And this is why the results of MultiResCNN are better than other models. For LD-PLAM, the replacement of CNN with bidirectional LSTM can automatically capture the keyword information of different lengths by adjusting the gate units. In summary, on both Chinese Xiangya full and Xiangya 50 datasets, LD-PLAM reaches the best on all of the indicators, with MiF 0.806 (Xiangya full), MiAUC 0.994 (Xiangya full), P@8 0.413 (Xiangya full), MiF 0.856 (Xiangya 50), MiAUC 0.991 (Xiangya 50), P@5 0.477 (Xiangya 50) which are 8.5\%, 2.7\%, 9.3\%, 4.8\%, 0.4\%, 1.7\% higher than MultiResCNN, respectively. 

\begin{table}[ht]
\caption{Comparison of LD-PLAM and other competing models on Xiangya dataset. }
\centering
\setlength{\tabcolsep}{1mm}{
\begin{tabular}{ccccccccccccc}
\toprule
\multirow{2}{*}{Models} &  \multicolumn{3}{c}{Xiangya full} & \multicolumn{3}{c}{Xiangya 50}\\
\cmidrule(lr){2-4}\cmidrule(lr){5-7}
                        &   MiF   &  MiAUC  &   P@8    &   MiF   &  MiAUC  &   P@5   \\
\midrule
DeepLabeler             &  0.607  &  0.964  &  0.333   &  0.751  &  0.984  &  0.457  \\
KAICD                   &  0.654  &  0.954  &  0.344   &  0.786  &  0.988  &  0.465  \\
CAML                    &  0.730  &  0.972  &  0.372   &  0.803  &  0.985  &  0.463  \\
DR-CAML                 &  0.709  &  0.954  &  0.357   &  0.804  &  0.982  &  0.460  \\
MVC-LDA                 &  0.711  &  0.968  &  0.371   &  0.798  &  0.984  &  0.466  \\
MVC-RLDA                &  0.697  &  0.968  &  0.368   &  0.797  &  0.984  &  0.467  \\
MultiResCNN             &  0.743  &  0.968  &  0.378   &  0.817  &  0.987  &  0.469  \\
\midrule
LD-PLAM                 &  \textbf{0.806}  &  \textbf{0.994}  &  \textbf{0.413}   &  \textbf{0.856}  &  \textbf{0.991}  &  \textbf{0.477}  \\
\bottomrule
\end{tabular}
}
\label{tab:2}
\end{table}

\subsection{Prediction of New ICD Codes}

Predicting new ICD codes is a big challenge and extremely important in this field. In this study, LD-PLAM is applied to predict rare or new ICD codes. Following Anthony et al. \cite{Anthony2019Few}, on MIMIC-III full dataset, we split the ICD codes into three groups S (seen ICD codes), F (few-shot ICD codes) and Z (zero-shot ICD codes). Group S contains ICD codes that occur more than 5 times in training set. Group F contains ICD codes that occur between 1 and 5 times. Group Z contains ICD codes that never occur in the training set, but occur in the validation/testing set. R@10 is defined as how many real labels are recalled at the most possible 10 labels predicted by the models, which is used as the evaluation metric. 

For the baselines, DeepLabeler, KAICD, (DR-)CAML, MVC-(R)LDA and MultiResCNN cannot predict new ICD codes because these models focus on the learning of EMR on fixed classification, rather than the learning of the relationship between EMR and ICD codes. For ESZSL and ZAGCNN, ESZSL-W2V, ESZSL-W2V2, ESZSL-GRALS, ZACNN, ZAGCNN are obtained through different experimental settings (see \cite{Anthony2019Few} for more details). As shown in Table \ref{tab:3}, our model generally outperforms other competing models in R@10 on the three groups. Comparing with the previous optimal method ZAGCNN, LD-PLAM increase R@10 by 4.7\%, 37.0\% and 11.7\% in group S, F, and Z, respectively. It indicates that RoBERTa expresses the ICD codes with high quality, and LD-PLAM reasonably predicts the relationship between EMR vectors and ICD vectors. In particular, LD-PLAM reached R@10 of 0.553 on group Z, which indicates that our model recalls more than half of the real ICD codes in the top 10 predicted ICD codes, which further proves the predictive ability of LD-PLAM on new ICD codes. As for R@5, LD-PLAM has a mediocre performance, which indicates that LD-PLAM may not be able to recall the positive ICD codes very fine-grained. Further precise reordering could improve this problem. 

\begin{table}[ht]
\caption{MIMIC-III results across frequent (S), few-shot (F), and zero-shot (Z) groups. }
\centering
\setlength{\tabcolsep}{1mm}{
\begin{tabular}{ccccccc}
\toprule
\multirow{2}{*}{Models} &  \multicolumn{2}{c}{S} & \multicolumn{2}{c}{F} & \multicolumn{2}{c}{Z}\\
\cmidrule(lr){2-3}\cmidrule(lr){4-5}\cmidrule(lr){6-7}
                        &   R@5   &  R@10 &   R@5    &   R@10   &  R@5  &   R@10    \\
\midrule
ESZSL+W2V               &  0.135  &  0.191  &  0.031   &  0.051  &  0.157  &  0.257   \\
ESZSL+W2V2              &  0.127  &  0.189  &  0.031   &  0.048  &  0.148  &  0.305   \\
ESZSL+GRALS             &  0.256  &  0.393  &  0.033   &  0.060  &  0.076  &  0.138   \\
ZACNN                   &  0.278  &  0.435  &  0.152   &  0.195  &  0.364  &  0.442   \\
ZAGCNN                  &  0.283  &  0.445  &  0.166   &  0.216  &  \textbf{0.428}  &  0.495   \\
\midrule
LD-PLAM                 &  \textbf{0.289}  &  \textbf{0.466}  &  \textbf{0.234}   &  \textbf{0.296}  &  0.426  &  \textbf{0.553}   \\

\bottomrule
\end{tabular}
}
\label{tab:3}
\end{table}


\subsection{Memory Analysis}
\begin{figure}
    \centering
    \includegraphics[scale=0.6, trim=10 0 0 0]{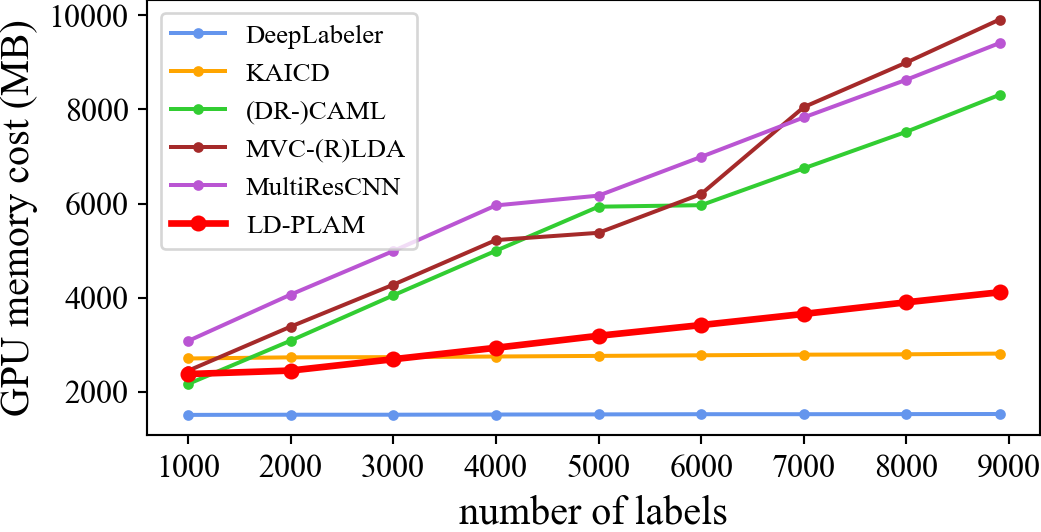}
    \caption{Curves of GPU cost with the number of labels for different models. }
    \label{fig:02}
\end{figure}

As mentioned before, pseudo label-wise attention has not only improved prediction accuracy, but also reduced GPU memory consumption. We record the GPU memory consumption of different number of labels (e.g. ICD codes) on MIMIC-III dataset when training different models in batch size 16. The curves of GPU cost with the number of labels are shown in Figure \ref{fig:02}. As we can see, the memory cost of models without label-wise attention such as DeepLabeler and KAICD have hardly improved with the increase of the number of labels. This is because we only need to add more neurons in output layer to predict more ICD codes, which requires very little memory. However, these models perform very poorly in predicting ICD codes if the number of codes is too large. For (DR-)CAML, MVC-(R)LDA and MultiResCNN, which reply on the label-wise attention modules, the memory consumption have improved greatly with the increase of the number of labels. About 1000 additional labels will increase the memory by 1000 MB. Although these models perform well in multi-label classification, excessive GPU memory consumption will lead high training cost. It is worth noting that these data are recorded when the text length is 2500 and the training batch size is only 16. As we know, a batch size of 16 is already a fairly small value, and too small a batch is likely to cause instability and even non-convergence of training \cite{DBLP:conf/nips/HeLT19}. With the advantages of pseudo label-wise attention, LD-PLAM, by contrast, has a relatively flat curve, and the GPU memory cost increases by only 200 MB for approximately 1000 more labels. As a result, on MIMIC-III full dataset, when the training batch size is set to 128 and the text length is 2500, we can use a single piece of 24GB graphics card to complete training, while we need to use a single piece of 80GB graphics card to complete the training of label-wise attention model. More importantly, despite the reduction in memory consumption, the performance of LD-PLAM is better. As shown in Table \ref{tab:1} and Table \ref{tab:2} above, the MiF value of LD-PLAM improves by 5.6\% and 7.1\% respectively in MIMIC-III full and Xiangya full datasets, compared to the best label-wise attention model (e.g. MultiResCNN). This shows that it is very necessary to use a pseudo label-wise attention mechanism to merge labels automatically and reduce computational redundancy in multi-label classification task, which can not only reduce the memory cost, but also improve the prediction performance. 

\subsection{Ablative Analysis}
\begin{figure}[ht]
    \centering
    \includegraphics[scale=0.4, trim=20 0 0 0]{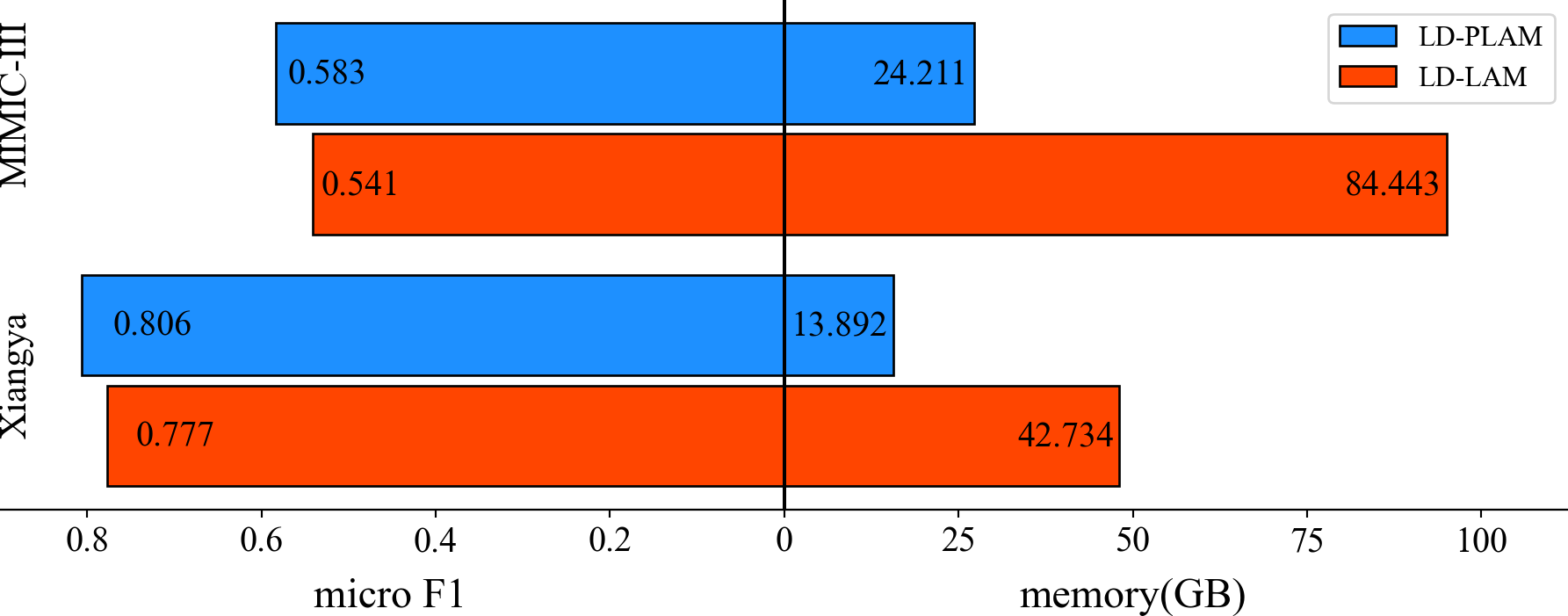}
    \caption{Ablative experiments about pseudo label-wise attention. }
    \label{fig:03}
\end{figure}

In order to further verify the effectiveness of pseudo label-wise attention, we remove it from LD-PLAM and obtain the model named LD-LAM. The comparisons of them on MIMIC-III full and Xiangya full datasets are shown in Figure \ref{fig:03}. As we can see, the MiF values of LD-PLAM in both MIMIC-III full and Xiangya full datasets are significantly higher then that of LD-LAM. The introduction of the pseudo label-wise attention has improved the results of the two datasets by 7.8\% and 3.7\%, respectively, and also reduced the GPU memory consumption by 71.3\% and 67.5\%, respectively. It is worth noting that: on MIMIC-III full dataset, according to our experimental settings, LD-LAM's GPU memory cost has reached to 84.4 GB due to the excessive number of labels (8,907 ICD codes) and large batch size (128 EMRs per batch); on Xiangya full dataset, LD-LAM has also reached to 42.7 GB. This means we need at least three pieces of NVIDIA V100 32GB to train the models. In contrast, LD-PLAM needs only 24.2 GB and 13.9 GB, and we can use only one NVIDIA V100 32GB to complete training. This further shows that the pseudo label-wise attention can not only reduce memory cost but also improve the prediction results. 

\begin{figure}[ht]
    \centering
    \includegraphics[scale=0.6, trim=0 0 0 0]{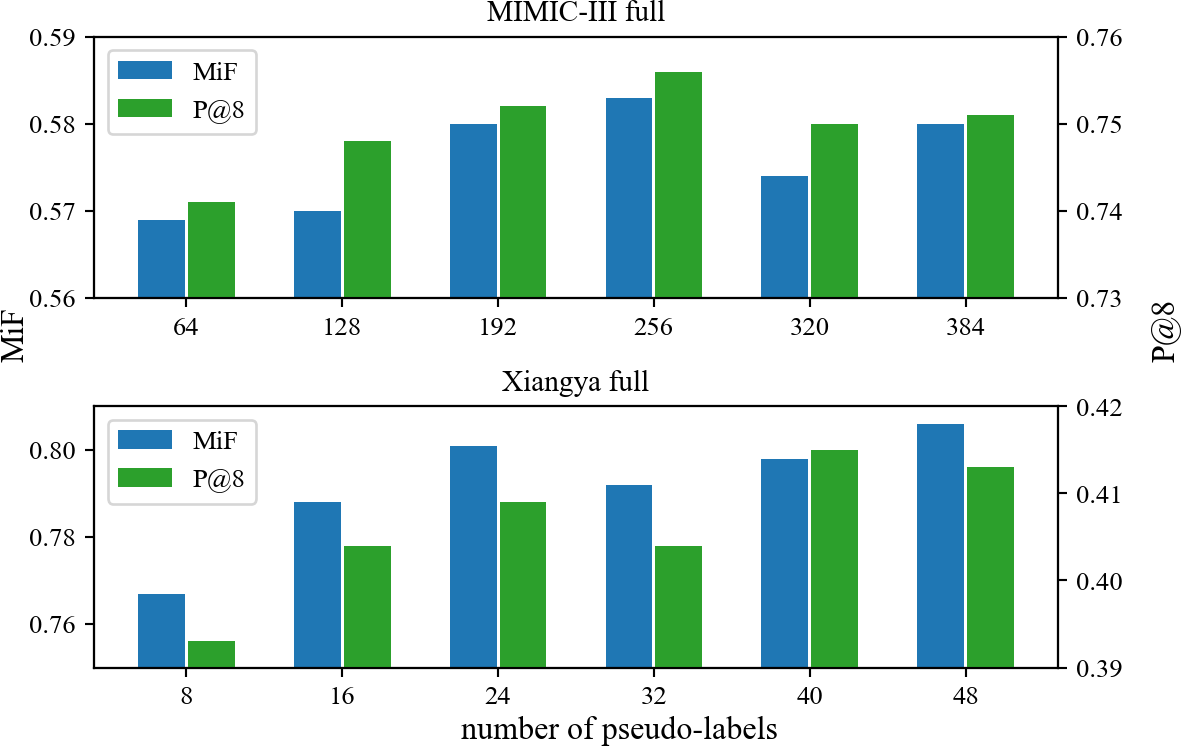}
    \caption{Effect of the number of pseudo labels. }
    \label{fig:07}
\end{figure}

The previous ablative experiment confirms the effectiveness of the pseudo label-wise attention, then a new question is raised: how should the number of pseudo labels be set? Therefore, we test the performance of LD-PLAM in MIMIC-III full and Xiangya full datasets with different number of pseudo labels and the results are shown in Figure \ref{fig:07}. For one thing, too small number of pseudo labels will limit the performance of the model. For example, on Xiangya full dataset, when the number of pseudo label is only 8, the MiF and P@8 are only 0.767, 0.393, which are 4.8\%, 4.8\% lower than that when the number is 48. This is in line with our expectations because too small number of pseudo label means too few attention modes, and too few attention modes are insufficient to adequately extract semantic features from different perspectives. At the extreme, when the number of pseudo labels is 1, it is equivalent to the models without label-wise attention like DeepLabeler, KAICD, etc. For another, when the number of pseudo labels is large enough, increasing the number will not improve the results. This means that we have already captured enough semantic information from enough perspectives, and increasing the number of attention modes only adds computational redundancy and overhead. This also confirms that the label-wise attention does have a lot of computational redundancy and it's not necessary to compute all these attention modes. As a result, we choose the number of pseudo labels to 256 for MIMIC-III full dataset, and 48 for Xiangya full dataset. 

\section{Discussion}

\subsection{Effect of label frequency}
For multi-label classification task, label frequency greatly affects the prediction performance of the label, in which the prediction performance of high-frequency label is higher than that of the low-frequency label. In order to figure out exactly which frequency LD-PLAM is improving, we analyze the prediction performance of different models at different label frequencies on Xiangya full dataset. According to the occurrence frequency of ICD codes in training set, we divide them into 11 sets: less than 10 times, 10 to 20 times, 20 to 30 times, ..., 90 to 100 times, more than 100 times. We choose area under the precision-recall curve (AUPR) to measure the prediction performance of different ICD code sets because AUPR can better evaluate unbalanced data. 

\begin{figure*}[ht]
    \centering
    \includegraphics[scale=0.45, trim=150 0 0 0]{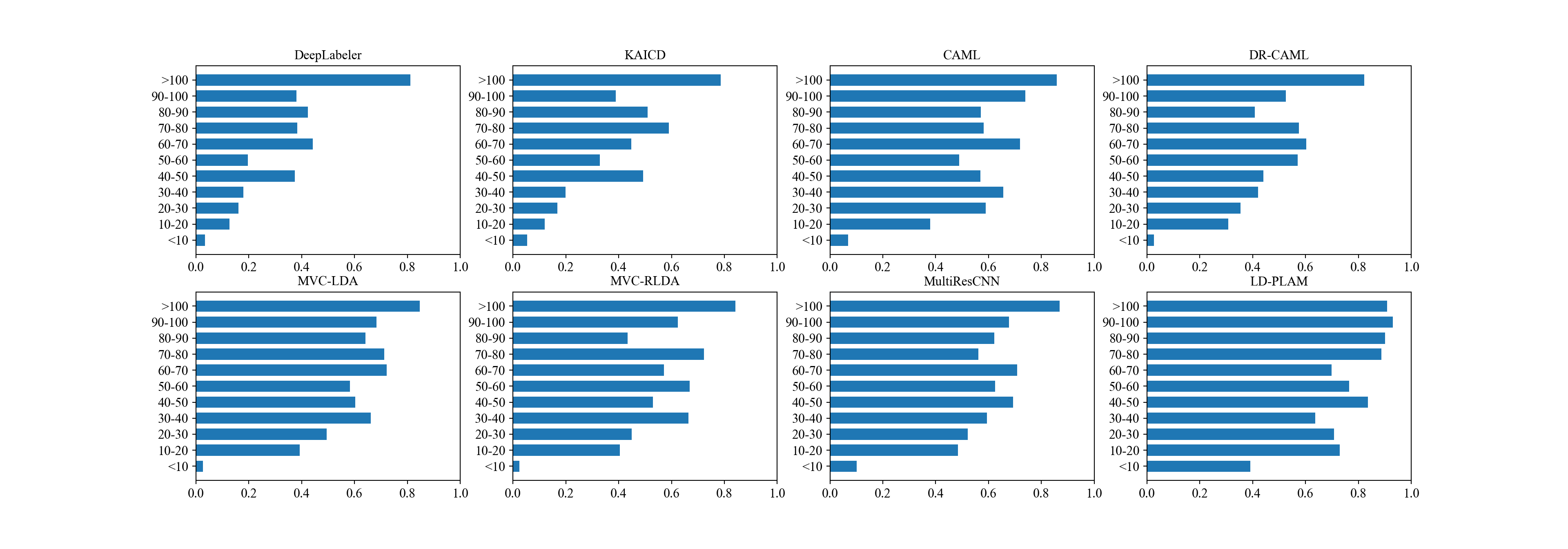}
    \caption{AUPR of different models at different label frequencies: in each sub figure, the horizontal axis represents the AUPR value, and the vertical axis represents the label frequency. }
    \label{fig:04}
\end{figure*}

As shown in Figure \ref{fig:04}, different sub figures represent different models: the horizontal axis is the AUPR value, the vertical axis is the label frequency. In line with our expectations, the general trend of different models is that the AUPR of high-frequency ICD codes is better than that of low-frequency ICD codes. DeepLabeler and KAICD predict an AUPR higher than 0.8 for ICD codes with frequencies higher than 100, while the AUPR of ICD codes appearing less than 40 times is even less than 0.2. With label-wise attention, CAML, DR-CAML, MVC-LDA, MVC-RLDA, MultiResCNN has improved the prediction performance of low-frequency ICD to a certain extent. The AUPR of ICD codes which occur between 10-100 times generally exceeded 0.5, while the codes with the occurrence frequency less than 10 have an AUPR of less than 0.1. This suggests that these models are quite unreliable for predicting extremely low frequency ICD codes. Relatively better, LD-PLAM increases the AUPR of ICD codes which appear less than 10 times to about 0.4, and also improves the AUPR of ICD codes with other frequencies. Compared with other models, the main improvement of LD-PLAM lies in the low frequency ICD codes, which is mainly due to the efficient pseudo label-wise attention and high-quality ICD vectors. This also confirms that pseudo label-wise attention is a more robust attention mechanism than label-wise attention. 

\subsection{Model interpretability}

\begin{table*}
\caption{The most critical words corresponding to some ICD codes. For the frequent ICD codes, we present the four most critical words for each ICD code. For the rare ICD codes, there is only one EMR for each ICD code and thus only one most critical word is kept.}
\centering
\begin{tabular}{ccccccc}
\toprule
\multicolumn{2}{c}{ICD codes}         & Short title & Long title & Critical words \\
\midrule
\multirow{8}{*}{Frequent}  & 401.9& Hypertension NOS & Unspecified essential hypertension & \tabincell{c}{hypertension; htn; hypertensive; \\ lisinopril; ...}\\
                                & 272.4& Hyperlipidemia NEC/NOS & Other and unspecified hyperlipidemia & \tabincell{c}{hyperlipidemia; dyslipidemia; hld; \\ hypercholesterolemia; ...} \\
                                &427.31& Atrial fibrillation & Atrial fibrillation & \tabincell{c}{fibrillation; afib; fib; \\ atrial; ...} \\
                                & 428.0& CHF NOS & Congestive heart failure, unspecified & \tabincell{c}{chf; diastolic; congestive; \\ heart; ...} \\
                                &414.01& Crnry athrscl natve vssl & \tabincell{c}{Coronary atherosclerosis of \\ native coronary artery} & \tabincell{c}{cad; mammary; lima; \\ coronary; ...} \\
\midrule
\multirow{7}{*}{Rare} &825.31& Fx astragalus-open & Open fracture of astragalus & talus; \\
                                &379.41& Anisocoria & Anisocoria & anisocoria; \\
                                &304.00& Opioid dependence-unspec & Opioid type dependence, unspecified & methadone; \\
                                & 860.1& Traum pneumothorax-open & \tabincell{c}{Traumatic pneumothorax \\ with open wound into thorax} & pneumothorax; \\
                                &524.60& TMJ disorders NOS & \tabincell{c}{Temporomandibular joint disorders, \\ unspecified} & tmj; \\                         
\bottomrule
\end{tabular}
\label{tab:4}
\end{table*}

Interpretability has always been a challenge in deep learning, and the introduction of attention \cite{bahdanau2014neural} points out a direction for interpretability in deep learning. In automatic ICD coding, we should pay special attention to the interpretability in model's inference, which can not only improve the reliability of the model in assisting manual coding, but also let us more clearly understand our model. Assuming $l_r$ is the length of an EMR. for label-wise attention, we can take out an attention vector $\gamma_i \in \mathbb{R}^{(1 \times l_r)}$ corresponding to ICD code $i$ to observe the inference of a model. For pseudo label-wise attention, we can do a similar thing to extract the attention vector. The difference is that in pseudo label-wise attention, we will generate more than one attention vector for each ICD code during the computation. As mentioned in section 3.1-3.3, we can get attention vectors $\alpha_i^1 \in \mathbb{R}^{d_c \times l_r^{(1)}}, \alpha_i^2 \in \mathbb{R}^{l_r^{(1)} \times l_r^{(2)}}, ..., \alpha_i^K \in \mathbb{R}^{l_r^{(K-1)} \times l_r^{(K)}}, \beta_i \in \mathbb{R}^{l_r^{(K)} \times n}$. Among them, $l_r^{(k-1)}$ represents the number of pseudo ICD categories in $(k-1)$-th layer, $K$ is the number of pseudo label-wise attention layers, $\beta_i = \begin{pmatrix} \beta_{i1} & \beta_{i2} & ... & \beta_{im} \end{pmatrix}$ is the similarity between an EMR's $m$ vectors and the ICD $i$'s vector $v_i$, $m$ is the final number of pseudo labels. These attention vectors obtained by LD-PLAM need to do matrix multiplication to get a final attention vector $\gamma_i = \alpha_i^1 \alpha_i^2 ... \alpha_i^K \beta_i$. 

For the model trained on MIMIC-III full, we calculate the final attention vector $\gamma_i$ of each EMR for each ICD code on test set, and count the words with the highest attention value corresponding to each predicted positive ICD code. We extract five most frequent ICD codes and five rarest ICD codes from the statistical results, and the most critical words corresponding to the ICD codes are shown in Table \ref{tab:4}. It can be seen that LD-PLAM can find accurate keywords for both frequent ICD codes and rare ICD codes. Among them, the most critical words are the noun interpretation or abbreviation of the corresponding ICD codes. In addition, LD-PLAM can infer ICD codes based on some drug's name. For example, the critical words of code 401.9 contains "lisinopril" which is a special drug for treating hypertension. In summary, this shows that LD-PLAM does have a reliable inference process that can assist manual ICD coding.

\subsection{Case Study}
In order to figure out how pseudo label-wise attention works, we compare and analyze the attention modes of LD-PLAM and LD-LAM whose pseudo label-wise attention is removed. As mentioned before, pseudo label-wise attention automatically merges the attention modes of similar ICD codes. For ICD-10, if the codes share the same first three characters, they are considered similar codes. We randomly select the attention modes of ICD codes beginning with "C18", "D35", "I69" or "K81" to show the mechanism of pseudo label-wise attention. We use LD-PLAM and LD-LAM to perform predictions on some EMRs of Xiangya test set, and retain the results of attention modes of the selected ICD codes. Note that the attention modes can be represented as a vector $\in \mathbb{R}^{1 \times l_r}$ assigning weights to every word in EMR, and thus we can evaluate the similarities between attention modes by using Euclidean distance. For each label-wise attention mode computed by LD-LAM, we look for its nearest pseudo label-wise attention modes computed by LD-PLAM. In this way, we can find the relationship between pseudo labels and real labels to explain how it compresses. 

\begin{figure}
    \centering
    \includegraphics[scale=0.75, trim=0 30 0 0]{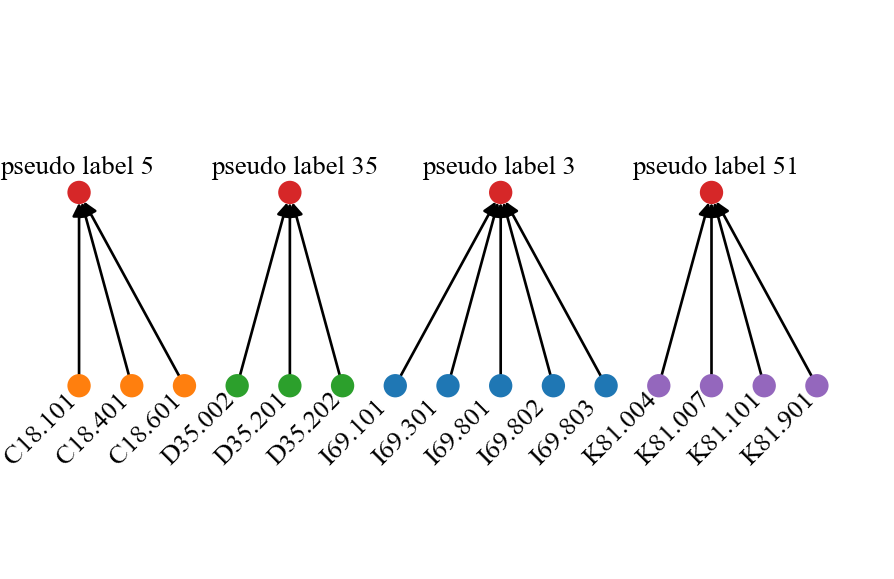}
    \caption{Illustration of 15 label-wise attention modes are compressed to 4 pseudo label-wise attention modes on Xiangya datasets. }
    \label{fig:05}
\end{figure}

As shown in Figure \ref{fig:05}, it is obvious that the similar ICD codes correspond to a same pseudo label. The attention modes of ICD codes beginning with "C18" correspond to the attention modes of "pseudo label 5"; ICD codes starting with "D35" correspond to the attention modes of "pseudo label 35"; ICD codes starting with "I69" correspond to the attention modes of "pseudo label 3"; ICD codes starting with "K81" correspond to the attention modes of "pseudo label 51". These ICD codes with the same prefix refer to similar diseases, and focused similar content of EMRs. Pseudo label-wise attention mechanism merges the 15 attention modes ("C18.101", etc.) into the 4 attention modes ("pseudo label 5", etc.), which indeed compresses the attention modes reasonably and reduces the computational redundancy.

\section{Conclusion}
Assisting manual ICD coding by deep learning model greatly improves the efficiency of ICD coding. General automatic ICD coding models relies on label-wise attention to improve the prediction performance. But at the same time, label-wise attention also greatly increase the computational cost and computational redundant due to a large number of ICD codes. In this paper, we propose a pseudo label-wise attention mechanism, which automatically merges the attention modes of similar labels. We apply the pseudo label-wise attention to automatic ICD coding, and our model achieves great performances on public MIMIC-III and private Xiangya datasets. Extensive experiments demonstrate the effectiveness of the pseudo label-wise attention. The case study shows how the pseudo label-wise attention merges similar ICD codes, which brings better generalization and efficiency. Furthermore, besides automatic ICD coding, pseudo label-wise attention should be applicable to other multi-label classification task because it is a general architecture which can be applied in many models.


%





\ifCLASSOPTIONcaptionsoff
  \newpage
\fi



\bibliographystyle{IEEEtran}
\bibliography{IEEEabrv.bib}

%



%
\newpage
\begin{IEEEbiography}[{\includegraphics[width=1in,height=1.25in,clip,keepaspectratio]{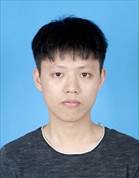}}]{Yifan Wu}
received his B.S. degree in computer science from Jishou University in 2019. He is currently a postgraduate student in the School of Computer Science and Engineering, Central South University, China. His currently research interests include bioinformatics, medical data mining and deep learning.
\end{IEEEbiography}
\begin{IEEEbiography}[{\includegraphics[width=1in,height=1.25in,clip,keepaspectratio]{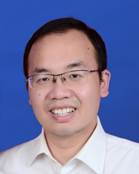}}]{Min Zeng}
received the B.S. degree from Lanzhou University in 2013, and the M.S. degree and Ph.D. degree in system science and computer science from Central South University in 2016 and 2020, respectively. He is currently a Lecturer in the School of Computer Science and Engineering, Central South University, Changsha, Hunan, P. R. China. His research interests include machine learning and deep learning techniques for bioinformatics and computational biology.
\end{IEEEbiography}
\begin{IEEEbiography}[{\includegraphics[width=1in,height=1.25in,clip,keepaspectratio]{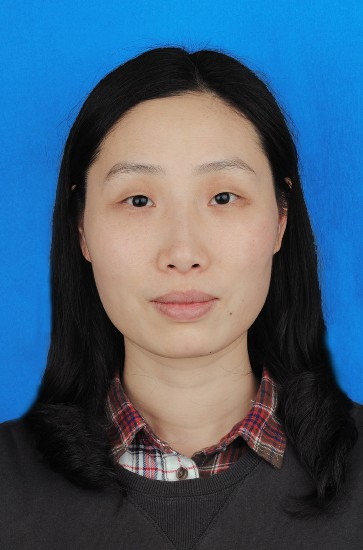}}]{Ying Yu}
received her B.S. degree and M.S. degree from University of South China in 2002 and 2009 separately. She is studying for a PhD in the School of Information Science and Engineering, Central South University, China. Her research focuses on machine learning, deep learning and analysis of healthcare big data.
\end{IEEEbiography}
\begin{IEEEbiography}[{\includegraphics[width=1in,height=1.25in,clip,keepaspectratio]{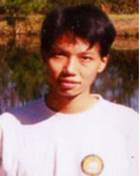}}]{Yaohang Li}
received the M.S. and Ph.D. degrees in computer science from Florida State University, Tallahassee, FL, USA, in 2000 and 2003, respectively. He is an Associate Professor in the Department of Computer Science at Old Dominion University, Norfolk, VA, USA. His research interests are in computational biology, Monte Carlo methods, and scientific computing. After graduation, he worked at Oak Ridge National Laboratory as a Research Associate for a short period. Before joining ODU, he was an Associate Professor in the Computer Science Department at North Carolina A\&T State University.
\end{IEEEbiography}
\begin{IEEEbiography}[{\includegraphics[width=1in,height=1.25in,clip,keepaspectratio]{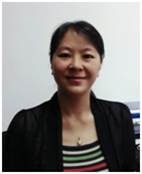}}]{Min Li}
received the PhD degree in Computer Science from Central South University, China, in 2008. She is currently a Professor at the School of Computer Science and Engineering, Central South University, Changsha, Hunan, P.R. China. Her research interests include computational biology, systems biology and bioinformatics. She has published more than 100 technical papers in refereed journals such as Bioinformatics, IEEE/ACM Transactions on Computational Biology and Bioinformatics, Proteomics, and conference proceedings such as BIBM, GIW and ISBRA. According to Google scholar, her paper citations are more than 6000 and H-index is 41.
\end{IEEEbiography}




\end{document}